# A Modular Dataset to Demonstrate LLM Abstraction Capability

Adam Atanas[1]     Kai Liu[1*]

[1]SES AI

## Abstract

Large language models (LLMs) exhibit impressive capabilities but struggle with reasoning errors due to hallucinations and flawed logic. To investigate their internal representations of reasoning, we introduce ArrangementPuzzle, a novel puzzle dataset with structured solutions and automated stepwise correctness verification. We trained a classifier model on LLM activations on this dataset and found that it achieved over 80% accuracy in predicting reasoning correctness, implying that LLMs internally distinguish between correct and incorrect reasoning steps, with the strongest representations in middle-late Transformer layers. Further analysis reveals that LLMs encode abstract reasoning concepts within the middle activation layers of the transformer architecture, distinguishing logical from semantic equivalence. These findings provide insights into LLM reasoning mechanisms and contribute to improving AI reliability and interpretability, thereby offering the possibility to manipulate and refine LLM reasoning.

## 1 Introduction

Recently, large language models (LLMs) based on the Transformer architecture (Vaswani et al., 2017) have demonstrated competence across a wide range of domains, from reading comprehension to coding to mathematics. However, in domains such as these, LLMs can often generate incorrect responses due to hallucinations and incorrect reasoning (Rawte et al., 2023). Large reasoning models (LRMs) explicitly trained to produce accurate chains of thought such as o1 (OpenAI et al., 2024) and DeepSeek-R1 (DeepSeek-AI et al., 2025) promise to increase the effectiveness of LLM reasoning. Even so, hallucinations and reasoning inaccuracies remain, preventing these models from exceling at more complex, multi-step tasks such as PlanBench (Valmeekam et al., 2023; Valmeekam et al., 2024) designed to evaluate the planning and reasoning capabilities of LLMs.

As LLM adoption increases and they begin to be used in increasingly critical applications, it becomes more necessary to detect and prevent such mistakes. The field of Explainable AI (reviewed in Ferrando et al., 2024) seeks to tackle this problem by trying to understand the inner workings of LLMs, and using that information to gain insight into how and why they go wrong.

For example, recent work suggests that LLMs understand the difference between truth and falsehood in factual statements (Azaria and Mitchell, 2023) internally. Specifically, the LLM contains representations of truthfulness that are strongest in its intermediate layers. Probing these layers for this representation can outperform directly prompting the LLM about the truthfulness of a statement (Liu et al., 2023). Interestingly, larger LLMs appear to have more capacity for abstraction, as their representations of truthfulness generalize better across different data modalities (Marks and Tegmark, 2024). Furthermore, understanding how truthfulness and other similar concepts are represented in the activations of LLMs can allow us to manipulate those representations to affect LLM behavior – for example, causing it to be more honest or less angry (Zou et al., 2023).

The recent advances in reasoning LLMs pose the analogous questions for reasoning: do LLMs have an internal concept of reasoning, and if so how much abstraction are they capable of in reasoning tasks? These are challenging questions to address with most publicly available reasoning datasets such as those used to train LRMs, which typically contain a variety of solutions with different structures that makes analysis difficult.

---

[*] Corresponding author: kai.liu.ai4science@gmail.com

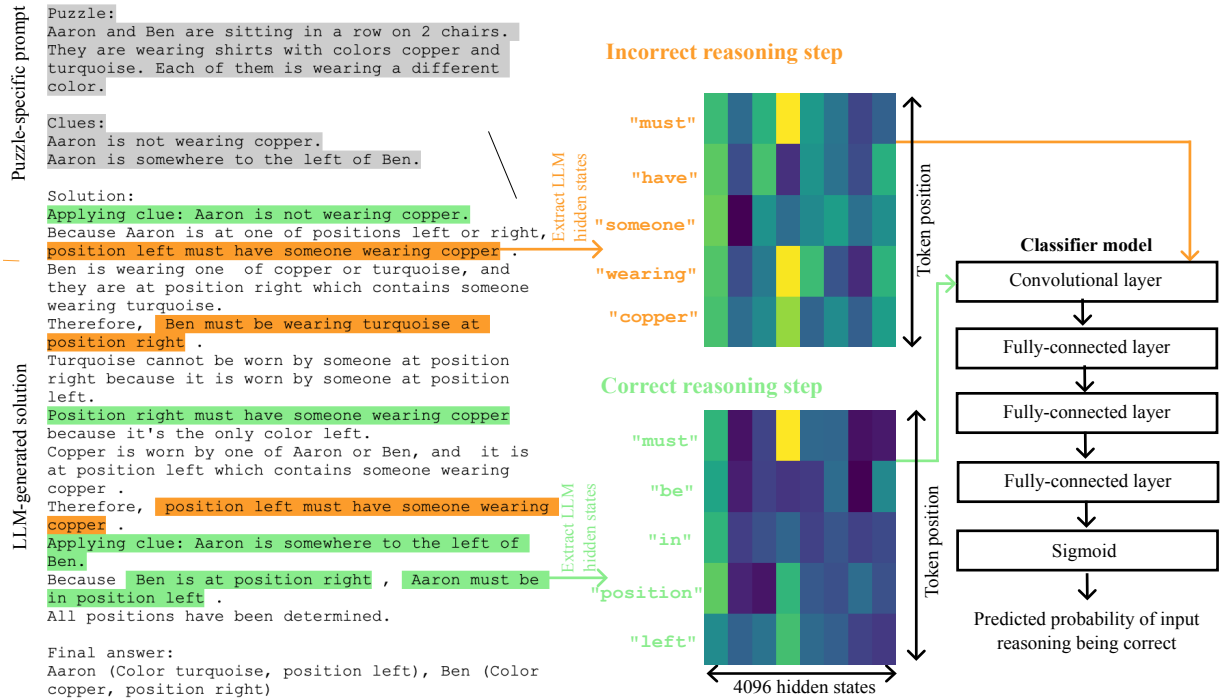

Figure 1: Diagram of how stepwise verification of LLM reasoning is used to train a classifier on the LLM's activations over those steps. The classifier is trained to predict whether a particular reasoning step is accurate.

In this paper, we introduce ArrangementPuzzle, a dataset suitable for analyzing LLM internal representations of reasoning. We use it to discover that LLMs have an innate understanding of when their reasoning steps are accurate, and they also have an internal representation of abstraction that separates logical from semantic equivalence.

## 2 ArrangementPuzzle Dataset

To test LLM reasoning representations, we constructed ArrangementPuzzle, a customizable puzzle dataset. This dataset and the code for generating it are publicly available in our GitHub repository https://github.com/Solid-Energy-Systems/arrangement_puzzle. Each puzzle contains some clues as to how a certain number of people are arranged, and what colors they are wearing. The LLM (Llama-3.1-8B-Instruct in our case) is then tasked with determining the full arrangement. An example puzzle is shown on the left of Figure 1.

The puzzles are guaranteed to have a unique solution given the available clues, and they are guaranteed to not have any redundant clues (that is, removing any single clue would result in a non-unique solution). Inspired by (Mirzadeh et al., 2024) and (Jiang et al., 2024), the puzzles randomize the exact names and colors used each time, as well as the correct arrangement and clues given. Our results differ from previous work in that our dataset allows statement-level (rather than solution-level) accuracy checking, and it contains a deterministic solution generator capable of generating full reasoning traces as it makes logical deductions to solve the puzzle (see Figure 5 for examples and Solution Generator for more details). Additionally, our focus is on using LLM activations derived from our dataset to understand LLM internal representations of reasoning, rather than performance benchmarking.

### 2.1 Reasoning Dataset

To this end, we evaluate the LLM on 10,000 prompts from puzzles with $n = 2$ people, of which it gets 67.6% correct. We save its activations for each generated token in each puzzle to disk. Simultaneously, we use a regular expression-based parser (also available on GitHub) to extract reasoning statements from the LLM's text output as it constructs partial claims about the correct arrangement (for example, `Alice is not sitting on the right`). This works because our prompting approach encourages the LLM to use very specific phrasing for its reasoning in line with the phrasing of the solution generator (see LLM Prompting).

Once the parser has extracted such statements, it evaluates them for correctness against the ground-truth solution to the puzzle. This then creates a dataset of model activations labeled by whether they came from a correct reasoning step or incorrect reasoning step.

## 3 Reasoning Classifier

We then use this dataset to train a classifier model (Figure 1). The model takes as input the LLM activations at a particular set of layers (typically 1 layer) at the last five token positions in a reasoning statement. It is trained to predict whether that reasoning statement was correct or incorrect. The classifier is a feedforward architecture that contains a single convolutional layer over token position followed by several fully-connected layers (more details in Classifier Architecture). It was trained on 7544 training puzzles and 1130 validation puzzles for 50 epochs and evaluated on a withheld set of 1141 test puzzles. The source code is available in our GitHub repository at https://github.com/Solid-Energy-Systems/activation_classifier.

### 3.1 Isomorphic puzzles

One important property of our dataset is that distinct puzzles can be generated by permuting some of the puzzle details (eg: names and colors). This, together with LLM output randomness, allow generating a variety of different data from a smaller set of distinct logic puzzles. However, to ensure our reasoning classifier learned information about actual logical reasoning and was not able to memorize patterns based on the exact clues given, we took steps to ensure that the validation and test datasets contained distinct sets of logical puzzles. To this end, we define two puzzles to be isomorphic if there exists a permutation of the clues, and substitutions of names and colors, to transform one puzzle into the other. Then we ensure our training, validation, and testing datasets contain disjoint isomorphism classes of puzzles.

### 3.2 Classifier Performance

The performance of the classifier as a function of the transformer layer it was trained on is shown in Figure 2. This high level of performance (>80% for most layers) demonstrates that the LLM does in fact contain distinct representations of correct and incorrect reasoning patterns. Additionally, these representations of reasoning appear to be strongest in the middle-late attention layers. This echoes previous findings (e.g. (Azaria and Mitchell, 2023)), which indicate that these layers also encode abstract representations of truth. We additionally ran an analysis where we trained a classifier on all of these top 5 performing layers (15, 17, 23, 25, and 30), but performance did not substantially increase (dashed red line), suggesting that these layers contain similar representations of the reasoning information.

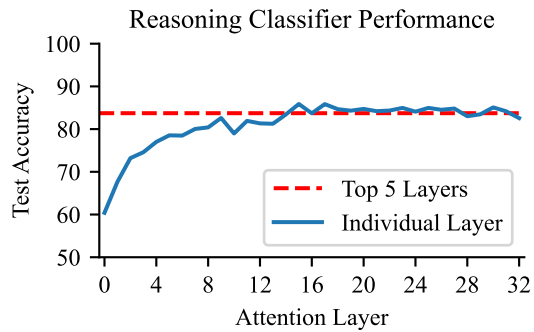

Figure 2: Performance of the reasoning classifier trained on a specific layer evaluated on testing data.

## 4 Reasoning Information Abstraction

Since our classifier is a neural network, it does not readily yield information on *how* the reasoning information is stored. However, it did reveal that the reasoning representation was strongest in the middle-late attention layers. Based on this, we hypothesized that these layers might store more abstract representations of reasoning and logic.

To test this hypothesis, we used our dataset to develop an abstraction test. Specifically, we algorithmically generated solutions for all our puzzles using our solution generator (see Solution Generator) and evaluated the LLM on these solutions and stored its layer activations. Notably, the LLM itself was not used to generate the text. We then used this dataset to compare the LLM's activations across solutions.

### 4.1 Information Abstraction in LLMs

Running the solution generator on our set of 10,000 puzzles, we identified two sets of puzzles of interest to studying abstraction in LLMs:

1. Logically distinct puzzles with solutions that contain identical lines of text, but at different places in the logical sequence.

2. Puzzles with isomorphic solutions – that is, where one solution can be transformed into the other by replacing details like names and colors.

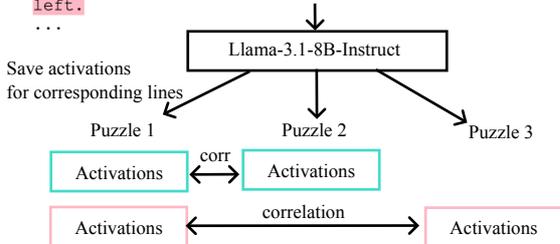

Figure 3: Examples of puzzles with distinct logical structure but identical text (Puzzles 1 and 2), and isomorphic puzzles with identical logical structure but distinct text (Puzzles 1 and 3). LLM activations on highlighted text are compared via correlation.

We randomly sampled 10,000 pairs of lines of text from each of these categories – that is, identical lines of text from logically distinct puzzles or corresponding but non-identical lines of text from isomorphic puzzle solutions (examples shown in Figure 3). For each pair of tokens in each pair of lines of text sampled in this way, we computed the correlation coefficient across hidden activations for a given layer between the two corresponding tokens. We then averaged this together across tokens in each line and then averaged it together across lines. We excluded pairs of lines with different numbers of tokens (eg: isomorphic lines where the token lengths of the substituted fields were not the same).

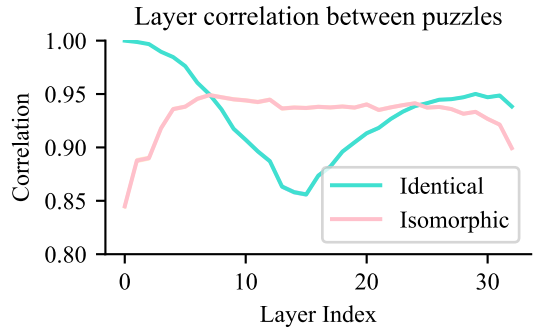

Figure 4: Correlation of attention layer activations between lines of text in different puzzles.

In this manner, we were able to compute the "abstraction level" of each layer in the Llama model by comparing the concordance of its activations between the two conditions (Figure 4). Layers with high correlation in the "Identical" condition and low correlation in the "Isomorphic" condition, such as the $0^{th}$ embedding layer, contain mostly token-specific information. On the other hand, layers where the reverse is true, such as layers 10-20, contain more abstract information about higher-level logical features of the puzzle.

## 5 Discussion

By leveraging ArrangementPuzzle, we trained a classifier that accurately distinguishes correct from incorrect reasoning steps, confirming that LLMs internally encode logical consistency. Specifically, our study demonstrates that LLMs possess internal representations of reasoning correctness, with the strongest signals emerging in middle-late Transformer layers. Additionally, our analysis of abstraction in model activations suggests that LLMs differentiate between logical and semantic equivalence. These findings contribute to a deeper understanding of LLM reasoning processes and may inform future efforts to enhance model reliability, interpretability, and trustworthiness. In particular, the fact that LLMs already encode an innate representation of reasoning may explain their ability to gain massive improvements in reasoning capability via distillation from LRMs, even without additional reinforcement learning (DeepSeek-AI et al., 2025) and even with as few as 1000 SFT samples for distillation (Muennighoff et al., 2025).

## 6 Limitations and Ethical Concerns

### 6.1 Limitations

While our study provides insights into the internal representations of reasoning in large language models (LLMs), it has several limitations. First, our analysis is restricted to a specific class of structured logic puzzles, which may not fully capture the complexity of reasoning required in more diverse real-world scenarios. The constrained nature of our dataset, where solutions follow deterministic patterns, may not generalize to open-ended reasoning tasks that require commonsense knowledge, probabilistic inference, or multi-modal reasoning. Additionally, we did not examine whether our classifier's success in distinguishing correct from incorrect reasoning steps generalizes to other types of reasoning problems, preventing us from making claims about the generalizability of the reasoning representations we uncovered.

Another key limitation lies in our reliance on probing techniques to analyze model activations. While we demonstrate that certain Transformer layers encode representations of reasoning correctness and abstraction, our approach does not provide a mechanistic explanation of how these representations emerge or how they influence downstream reasoning behavior. Furthermore, our classifier is trained on activations from a single LLM architecture, and it remains unclear whether these findings generalize across different model families, sizes, or training paradigms. Future work should explore more diverse reasoning benchmarks, conduct broader cross-model analyses including LRM models, and develop methods for directly steering LLM reasoning based on these learned representations.

## 7 Ethical Concerns

We do not anticipate any immediate ethical impact arising from our work. However, our work does highlight the potential for LLMs to prioritize pattern-matching over accuracy, as we have demonstrated that LLMs have an internal representation of reasoning accuracy yet output incorrect reasoning anyway. Additionally, latent LLM abstraction capabilities highlighted in this work suggest that it may be relatively easy to "jailbreak" open-weight LLMs with a small amount of fine-tuning into producing potentially dangerous output.

## 8 Use of AI Assistants

We employed the use of AI assistants, primarily ChatGPT (versions 4o, o1, and o3-mini), to help generate some of the code and text of this manuscript. The authors have examined all output of these assistants to ensure accuracy.

## A Solution Generator

The algorithm begins by initializing all possible assignments for people, positions, and colors, effectively considering every permutation. For example, at the beginning it would initialize `Andrew` as being at one of positions `left` or `right`. It then iteratively applies the clues to eliminate invalid combinations, updating the sets of possibilities for each entity based on the constraints provided by the clues. Through constraint propagation, the algorithm refines these possibilities by intersecting sets and removing options when only one remains for a given entity. This iterative process continues until a unique solution is found, solving the puzzle. Importantly, every time the algorithm updates its internal possibilities, it outputs a reasoning step in text format.

## B LLM Prompting

To prompt the LLM to solve our puzzles, we use a 3-shot approach where we append the actual puzzle (to be solved) to three example puzzle/solution pairs. These solutions were generated by our solution generator. This primes the LLM to reason through the puzzles using similar logic. The full 3-shot prompt is available on our `arrangement_puzzle` GitHub repository in the file `prompt.txt`.

## C Classifier Architecture

Our classifier uses a convolutional layer with 128 output channels and kernel size 3 that convolves over the five token positions, treating each individual activation (of the 4096 hidden units) and each layer as a different channel. These outputs are then fed through three fully-connected layers with hidden sizes 256 and 128, before finally producing a single logit which is then passed through a sigmoid to produce the final prediction.

```
Puzzle:
Ava and Blake are sitting in a row on 2 chairs.
They are wearing shirts with colors red and pink.
Each of them is wearing a different color.
Clues:
The person wearing pink is not sitting on the far left.
Blake is sitting on the far right.
Solution:
Applying clue: The person wearing pink is not sitting on the far left.
The person wearing pink cannot be at position left.
Red cannot be worn by someone at position right because it is worn by someone at position left.
Position right must have someone wearing pink because it's the only color left.
Applying clue: Blake is sitting on the far right.
Blake must be at position right.
Ava cannot be at position right because they are at position left.
Position right must have Blake because they're the only person left.
Ava is wearing one of pink or red, and they are at position left which contains someone wearing red.
Therefore, Ava must be wearing red at position left.
Red is worn by one of Ava or Blake, and it is at position left which contains Ava.
Therefore, position left must have Ava wearing red.
Blake is wearing one of pink or red, and they are at position right which contains someone wearing pink.
Therefore, Blake must be wearing pink at position right.
Pink is worn by one of Ava or Blake, and it is at position right which contains Blake.
Therefore, position right must have Blake wearing pink.
All positions have been determined.

Final answer:
Ava (Color red, position left), Blake (Color pink, position right)
```

```
Puzzle:
Andrew and Bella are sitting in a row on 2 chairs. They are wearing shirts with colors mint and chocolate. Each of them is wearing a different color.
Clues:
The person wearing chocolate is not sitting on the far left.
Bella is sitting on the far right.
Solution:
Applying clue: The person wearing chocolate is not sitting on the far left.
The person wearing chocolate cannot be at position left.
Mint cannot be worn by someone at position right because it is worn by someone at position left.
Position right must have someone wearing chocolate because it's the only color left.
Applying clue: Bella is sitting on the far right.
Bella must be at position right.
Andrew cannot be at position right because they are at position left.
Position right must have Bella because they're the only person left.
Andrew is wearing one of chocolate or mint, and they are at position left which contains someone wearing mint.
Therefore, Andrew must be wearing mint at position left.
Mint is worn by one of Andrew or Bella, and it is at position left which contains Andrew.
Therefore, position left must have Andrew wearing mint.
Bella is wearing one of chocolate or mint, and they are at position right which contains someone wearing chocolate.
Therefore, Bella must be wearing chocolate at position right.
Chocolate is worn by one of Andrew or Bella, and it is at position right which contains Bella.
Therefore, position right must have Bella wearing chocolate.
All positions have been determined.

Final answer:
Andrew (Color mint, position left), Bella (Color chocolate, position right)
```

```
Puzzle:
Aaron and Blake are sitting in a row on 2 chairs. They are wearing shirts with colors mint and lilac. Each of them is wearing a different color.
Clues:
Blake is somewhere to the right of the person wearing mint.
Solution:
Applying clue: Blake is somewhere to the right of the person wearing mint.
Because the person wearing mint is at one of positions left or right, Blake must be in position right.
Because Blake is at position right, the person wearing mint must be in position left.
Aaron cannot be at position right because they are at position left.
Position right must have Blake because they're the only person left.
Blake is wearing one of lilac or mint, and they are at position right which contains someone wearing lilac.
Therefore, Blake must be wearing lilac at position right.
Lilac cannot be worn by someone at position left because it is worn by someone at position right.
Position left must have someone wearing mint because it's the only color left.
Lilac is worn by one of Aaron or Blake, and it is at position right which contains Blake.
Therefore, position right must have Blake wearing lilac.
All positions have been determined.

Final answer:
Aaron (Color mint, position left), Blake (Color lilac, position right)
```

Figure 5: Example puzzles with solutions from our deterministic solution generator. The left two puzzles are isomorphic to each other.